\documentclass[conference]{IEEEtran}
	\input Macros/alphabet
	\input Macros/abrege
\def\pth#1{\left(#1\right)}                
              
\def\cro#1{\left[#1\right]}                
               
\def\norm#1{\left\|#1\right\|}

\def\Exp#1{\exp\cro{#1}}

\def\AND{\text{and\:}}

\newsavebox{\fminibox}
\newlength{\fminilength}
\newenvironment{fminipage}[1][\linewidth]
  {\setlength{\fminilength}{#1}
   \begin{lrbox}{\fminibox}\begin{minipage}{\fminilength}}
  {\end{minipage}\end{lrbox}\noindent\fbox{\usebox{\fminibox}}}

	\def\pmu{^{-1}}

   \def\wh#1{\widehat{#1}}                 

	\def\T{^\tD} 
	\def\I{\,|\,}           

   \def\rond#1{\overset{\kern-0.33em~_\circ}{#1}}
   \def\rondit[#1]#2{\overset{\kern#1~_\circ}{#2}}

\def\babs{\begin{abstract}}             \def\eabs{\end{abstract}}
\def\barr{\begin{array}}                \def\earr{\end{array}}
\def\bcc{\begin{center}}                \def\ecc{\end{center}}
\def\bdes{\begin{description}}          \def\edes{\end{description}}
\def\bdoc{
\begin{document}}             \def\edoc{\end{document}}
\def\ben{\begin{enumerate}}             \def\een{\end{enumerate}}
\def\barr{\begin{array}}                \def\earr{\end{array}}
\def\beqn{\begin{eqnarray}}             \def\eeqn{\end{eqnarray}}
\def\beqnx{\begin{eqnarray*}}           \def\eeqnx{\end{eqnarray*}}
\def\bseqn{\begin{subeqnarray}}         \def\eseqn{\end{subeqnarray}}
\def\beq#1\eeq{\begin{equation}#1   	\end{equation}}
\def\bal#1\eal{\begin{align}#1\end{align}}
\def\balx#1\ealx{\begin{align*}#1\end{align*}}
\def\beqx{$$}                           \def\eeqx{$$}
\def\bfig{\protect\begin{figure}}       \def\efig{\protect\end{figure}}
\def\bfigx{\protect\begin{figure*}}     \def\efigx{\protect\end{figure*}}
\def\bfigt{\protect\begin{figurette}}   \def\efigt{\protect\end{figurette}}
\def\bfl{\begin{flushleft}}             \def\efl{\end{flushleft}}
\def\bfr{\begin{flushright}}            \def\efr{\end{flushright}}
\def\bit{\begin{itemize}}               \def\eit{\end{itemize}}
\def\bmi{\begin{minipage}}              \def\emi{\end{minipage}}
\def\bfmi{\begin{fminipage}}            \def\efmi{\end{fminipage}}
\def\bpic{\begin{picture}}              \def\epic{\end{picture}}
\def\bqun{\begin{quotation}}            \def\equn{\end{quotation}}
\def\bsl{\begin{slide}}               	\def\esl{\end{slide}}
\def\btabb{\begin{tabbing}}             \def\etabb{\end{tabbing}}
\def\btabl{\begin{table}}               \def\etabl{\end{table}}
\def\btablx{\begin{table*}}             \def\etablx{\end{table*}}
\def\btabu{\begin{tabular}}             \def\etabu{\end{tabular}}
\def\btabx{\begin{tabular*}}            \def\etabx{\end{tabular*}}
\def\bbib{\begin{thebibliography}{999}} \def\ebib{\end{thebibliography}}
\def\bver{\begin{verbatim}}             \def\ever{\end{verbatim}}
\def\bca{\begin{cases}}                 \def\eca{\end{cases}}
\def\brk{\begin{remark}~}                \def\erk{\end{remark}}

\def\cl#1{\centerline{#1}}

    \usepackage{cite}
    \usepackage{amsmath,amssymb,amsfonts}
    \usepackage{graphicx}
    \usepackage{xcolor}
    \usepackage{verbatim}
	\usepackage{amsthm}
   	\usepackage{tikz}\usetikzlibrary{shapes.geometric,patterns,arrows}
   	%

\newtheoremstyle{TheRemark}
    {5pt}
    {5pt}
    {\normalshape}
    {0pt}
    {\itshape}
    {~---}
    {0ex}
    {\thmname{#1}\thmnumber{ #2}\thmnote{~(\textbf{#3~})}}

\theoremstyle{TheRemark}
\newtheorem{remark}{Remark}

	\definecolor{Bleu}{rgb}{0,0,0.75}    \def\Blue#1{{\color{Bleu} #1}}
	\definecolor{Rouge}{rgb}{0.75,0,0}   \def\Red#1{{\color{Rouge} #1}}
	\definecolor{Vert}{rgb}{0,0.7,0}     \def\Green#1{{\color{Vert} #1}}
    \definecolor{Marron}{rgb}{0.5,0.5,0} \def\Brown#1{{\color{Marron} #1}}
    \definecolor{Magenta}{rgb}{1,0,1}    \def\Magenta#1{{\color{Magenta} #1}}
    \definecolor{Gris}{rgb}{1,0,1}    	\def\Gray#1{{\color{gray} #1}}

	\def\fns{\footnotesize}
	\def\sss{\scriptscriptstyle}
	\def\For{\xb}
	\def\Back{\xb}
    \def\VarFor{v^{\sss +}}		\def\SigFor{\sigma^{\sss +}}
    \def\VarBack{v^{\sss -}}	\def\SigBack{\sigma^{\sss -}}
	\def\colog{\mathrm{colog}\,}
    \usepackage{pifont} \def\MyCheck{\ding{51}}
	\def\Quote#1{``#1''}

\bdoc

\title{Estimation of instrument and noise parameters \\ for inverse problem based on prior diffusion model}
 
\author{
\IEEEauthorblockN{Jean-Fran\c{c}ois~Giovannelli}
\IEEEauthorblockA{\textit{Groupe Signal-Image, IMS (Univ. Bordeaux, CNRS, BINP), Talence, France} }
%
%
%
}

\maketitle

\begin{abstract}
%
%
This article addresses the issue of estimating observation parameters (response and error parameters) in inverse problems. The focus is on cases where regularization is introduced in a Bayesian framework and the prior is modeled by a diffusion process. In this context, the issue of posterior sampling is known to be thorny, and a recent paper~\cite{Giovannelli26} proposes a notably simple and effective solution. Additionally, it opens an remarkable flexibility when it comes to estimating observation parameters. The proposed strategy enables to define an optimal estimator for both observation parameters and image of interest. Furthermore, the strategy provides a means for uncertainty quantification. In addition, MCMC algorithms allow for the computation of estimates and properties of posteriors, while offering some guarantees. The paper presents several numerical experiments that clearly confirm the computational efficiency and the quality of both estimates and uncertainty quantification. 
\end{abstract}


\section{Introduction and problem formulation} \label{Sec:Intro}

%
%
The present paper deals with the resolution of inverse problems~\cite{Hansen10,Kaipio05,Santamarina05,Vogel02} when the observation system is modeled by a linear operator and an additive Gaussian error:
\beq\label{Eq:ModelDirect}
\yb = \Hb_\iotab \xb_{\sss 0} + \eb,
\eeq
where $\xb_{\sss 0}\in\eR^P$ collects the unknowns, $\yb,\eb\in\eR^M$ collect the measurements and errors, and $\Hb_\iotab\in\eR^P\times \eR^M$ characterizes the operator, \eg a convolution.
The vector $\iotab$ parametrizes the instrument response, typically the width of a Lorentz Point Spread Function (PSF). This is one of the key parameters to be estimated. The second one, denoted $\etab$, controls the error pdf, \eg mean $m_e$ and variance $v_e$ of an homogeneous white noise.
%
%
All these parameters are gathered in the vector $\thetab=\cro{\iotab,\etab}$. These parameters are included among the unknowns and this is a crucial feature of the proposed method to estimate them in addition to the image of interest.
%

%
%
%

The ability to estimate observation parameters, in addition to the image of interest, is crucial in practice. It is common to have information on instrument parameters or noise levels, \eg a nominal value with an associated uncertainty, but it is rare to know them exactly. Moreover, failing to account for uncertainties in these parameters leads to erroneous uncertaintiy quantification about the image of interest.

This issue has been frequently addressed and several solutions have been proposed~\cite{Yan26,Pereyra13,Orieux12,Orieux10,Dobigeon09,Bishop08,Campisi07,Mugnier04} 
referred to as auto-adjusted, adaptive, self-tuned, myopic/blind or self-calibrated\dots
That said, in the case of priors constructed from the recent diffusion models~\cite{Chan24,Ribeiro25,Nakkiran25}, this issue remains difficult and has been very little addressed (however, see~\cite{Murata23} and Remark\,\ref{RQ:GibbsDDRM}). 
The difficulty may be due to the fact that the dominant approaches are inherited from ancestral sampling (designed for the prior): they attempt to correct the latter to produce posterior samples. But whilst they are exact for the prior they are approximated for the posterior. For example, to sample the posterior for the images, \cite{Chung24} and \cite{Song23} rely on approximations that involve $\Hb_\iotab$ itself and this complicates or even makes it impossible to manage the parameter~$\iotab$. 
In contrast, G-DSP (Gibbs Diffusion Posterior Sampling) recently proposed~\cite{Giovannelli26} clearly takes advantage of the Markov structure and conditional independences (see also Fig.\,\ref{Fig:Hierarchy}), which opens up noticeable possibilities for the estimation of observation parameters and gives raise to the present contribution, referred to as Hyper-G-DPS (Hyperparameter-G-DPS).

The paper is organized as follows. Section\,\ref{Sec:Posterior} introduces the  various distributions to model measurements and unknown quantities. Section \,\ref{Sec:PosteriorSampler} describes the posterior sampler. The numerical assessement using the MNIST example set is given in Section\,\ref{Sec:Simulations}. Finally, Section\,\ref{Sec:Conclusion} proposes a synthesis and includes a few perspectives. Part of the calculations are reported in the Appendix.

\section{Likelihood, prior, posterior}\label{Sec:Posterior}

\subsection{Noise and measurement}

The measurements are included \via the likelihood deduced from the observation model~(\ref{Eq:ModelDirect}) and a model for the error. Here, the latter is described as Gauss with mean $\mb_e$ and covariance $\Cb_e$ that is to say $\Nc(\eb \,;\, \mb_e, \Cb_e$). So, the likelihood of the unknowns $(\xb_{\sss 0},\thetab)$ attached to the measurement $\yb$ reads
\beq \label{Eq:Likelihood}
f(\yb  \I \xb_{\sss 0},\thetab)  = \Nc(\yb \,;\, \mb_e+\Hb_\iotab \,\xb_{\sss 0},\Cb_e) \,.
\eeq

Regarding the error, in subsequent developments, we focus on the stationary and white case: the mean and variance are homogeneous and denoted by $m_e$ and $v_e$ respectively and collected in the vector $\etab=\cro{m_e, v_e}$. However, the proposed methodology can easily be generalised to cover more complex situations, and could incorporate correlation parameters, or non-Gaussian noise based on location mixture of Gaussians.

Regarding the vector $\iotab$, it collects the parameters of the instrument response. It may include the amplitude and width of the PSF, \eg  a Lorentzian as considered in the numerical study (Sect.\,\ref{Sec:Results}). However, the proposed methodology can easily be generalised to more complex PSFs (see, \eg \cite{Yan23,Fetick20}). 



The vector $\thetab=\cro{\iotab,\etab}$ collects the unknown observation parameters (instrument and error), the other unknown being  the image of interest $\xb_{\sss 0}$. The aim of the rest of this section is to incorporate the available knowledge about these unknowns through probability distributions.
\bit
\item 
%
%
With regard to images, traditional approaches rely, for example, on pixel positivity, pixel correlation, contours or pulses\dots Here, we rely on the fact that the image shares a certain resemblance with available examples.
\item
%
%
Regarding the observation parameters, the available information may be an order of magnitude, a nominal value with uncertainty, a minimum\,/\,maximum values,\dots
\eit
%
%
When the available information is more uncertain, it is referred to as a poorly informative prior.

%
Among the distributions that allow this information to be taken into account, one seeks to assign a prior so that the posterior (and especially its conditionals) is easy to manipulate and sample. With this in mind, whenever possible, one relies on simple models, such as Gaussian models, and/or on the notion of conjugacy~\cite{Robert07}. 

\subsection{Prior for unknowns}\label{Sec:HyperInstru}

%
\smallskip\textsl{Observation parameter $\iotab$}~---~For the instrument parameter~$\iotab$, for each component, we define a uniform prior between a minimum and a maximum values  in line with the knowledge of the physical principles of the instrument. We simply write 
\beq \label{Eq:PriorInstru}
f_\Ib(\iotab) = \Uc(\iotab) \,.
\eeq
In the numerical study of Sect.\,\ref{Sec:Results} we will consider a Lorentz PSF and $\iota$ encode the width.

%
\smallskip\textsl{Noise parameter: offset $m_e$}~---~Regarding the level of offset in measurements, we consider a situation where a nominal value $m_0$ and a precision $p_0$ are available and we define
\beq \label{Eq:PriorNoiseOffset}
f_M(m_e) = \Nc(m_e \,;\, m_0, p_0\pmu)  \,.
\eeq
In the numerical study of Sect.\,\ref{Sec:Results}, we will consider the poorly informative case: $p_0$ is small (and $m_0=0$).

\smallskip\textsl{Noise parameter: scale $\gamma_e$}~---~Regarding $\gamma_e$ (for notational convenience $\gamma_e=1/v_e$), a classical choice is a Gamma pdf:
\beq\label{Eq:PriorNoiseScale}
f_\Gamma(\gamma) = \Gc(\gamma; a_0, b_0)
\eeq
This choice makes it easy to consider a nominal value with uncertainty based on the mean $a_0/b_0$ and the variance $a_0/b_0^2$. 
%
%
%

%
\smallskip\textsl{Diffusion prior for the images}~---~This prior is described using a diffusion model~\cite{Chan24,Ribeiro25,Nakkiran25}: essentially, available examples are transformed into noise, and conversely, new examples are generated by transforming noise realisations. To achieve this, the methodology consists in introducing (i) $T$ latent variables $\xb_{\sss 1:T}$ (in addition to $\xb_{\sss 0}$) and an extended prior $\pi_{\sss 0:T}(\xb_{\sss 0:T})$ and (ii)  two joint pdfs for $\xb_{\sss 0:T}$: a \textsl{forward} denoted $p^{\sss +}_{\sss 0:T}$ and a \textsl{backward} denoted $p^{\sss -}_{\sss 0:T}$. For practical efficiency, both are chosen in Markovian form:  
\beqn
p^{\sss +}_{\sss 0:T}(\For_{\sss 0:T})
 &=& p^{\sss +}_{\sss 0} (\For_{\sss 0}) \, \prod\nolimits_{t=1}^{T} \, p^{\sss +}_{\sss t|t-1} (\For_{\sss t} \I \For_{\sss t-1}) \label{Eq:Forward} \\[0.1cm]
p^{\sss -}_{\sss 0:T}(\Back_{\sss 0:T}) 
 &=& p^{\sss -}_{\sss T} (\Back_{\sss T}) \, \prod\nolimits_{t=1}^{T} p^{\sss -}_{\sss t-1|t} (\Back_{\sss t-1} \I \Back_{\sss t}) \label{Eq:Backward}
\eeqn
which involves two terminal marginal pdfs $p^{\sss +}_{\sss 0}$ and $p^{\sss -}_{\sss T}$ and two sets of transition pdfs $p^{\sss +}_{\sss t|t-1}$ and $p^{\sss -}_{\sss t-1|t}$. 
Regarding the terminals
\beqx
p^{\sss +}_{\sss 0} (\For_{\sss 0})  = \pi_{\sss 0}   (\For_{\sss 0}) ~~~\AND~~~
p^{\sss -}_{\sss T} (\Back_{\sss T}) = \Nc( \Back_{\sss T} \,;\, \zerob , \Ib)
\eeqx
the first is the pdf $\pi_{\sss 0}$ of the example set and the second is the pdf of noise (Gaussian, white and reduced). 
With regard to transitions, again for practical efficiency, Gaussians are chosen with the following parameters.
\beqn
p^{\sss +}_{\sss t|t-1} (\For_{\sss t}  \I \For_{\sss t-1})  &=& \Nc (\For_{\sss t}   \,;\, k_{\sss t} \, \For_{\sss t-1}	, \VarFor_{\sss t} \Ib  ) \label{Eq:ForwardTransit} \\[0.1cm]
p^{\sss -}_{\sss t-1|t} (\Back_{\sss t-1} \I \Back_{\sss t}) &=& \Nc ( \Back_{\sss t-1} \,;\, \mub_{\sss t}(\Back_{\sss t})	, \VarBack_{\sss t} \Ib ) \label{Eq:BackwardTransit}
\eeqn
The function $\mub_{\sss t}(\Back)$ is described by a neural network $\mub_{\sss t}^{\sss\pb}(\Back)$, with parameter $\pb$ and has two inputs: the image~$\Back$ and the time~$t$. Replacing $\mub_{\sss t}$ by $\mub^{\sss\pb}_{\sss t}$ in (\ref{Eq:BackwardTransit}), and substituting in (\ref{Eq:Backward}), yields $p^{\sss -,\pb}_{\sss 0:T}$.
The learning stage adjusts $\pb$ to minimise the Kullback distance between the forward $p^{\sss +}_{\sss 0:T}$ and the parametrized backward $p^{\sss -,\pb}_{\sss 0:T}$ pdfs while ensuring that the marginal pdfs for $\xb_{\sss 0}$ and $\xb_{\sss T}$ are
\beqx
    \pi_{\sss 0} =~ p^{\sss +}_{\sss 0} ~\simeq~ p^{\sss -}_{\sss 0} ~~~\AND~~~ \Nc	=~ p^{\sss -}_{\sss T} ~\simeq~ p^{\sss +}_{\sss T}
\eeqx	
\ie that of the example set and the noise. 
It suffices then to report the adjusted value of $\pb$ in $p^{\sss -,\pb}_{\sss 0:T}$ to obtain an adjusted joint backward pdf~(\ref{Eq:Backward}). Therefore, based on the latter, it is easy to sample the prior for $\xb_{\sss 0:T}$, starting from $t=T$ downto to $t=0$ and it is referred to as \textsl{ancestral sampling}. 

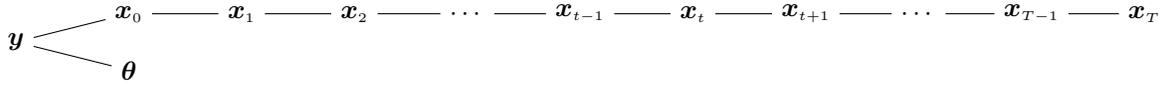
\begin{figure*}[ht]
\bcc
\begin{tikzpicture}[>=triangle 45,xscale=1,yscale=0.5,rotate=270]
\node{$\yb$} [grow=up]  
child{ node {$\thetab$}} 
child{ node {$\xb_{\sss 0}$} 
child{ node {$\xb_{\sss 1}$} 
child{ node {$\xb_{\sss 2}$} 
child{ node {$\cdots$} 
child{ node {$\xb_{\sss t-1}$} 
child{ node {$\xb_{\sss t}$} 
child{ node {$\xb_{\sss t+1}$} 
child{ node {$\cdots$} 
child{ node {$\xb_{\sss T-1}$} 
child{ node {$\xb_{\sss T}$} 
}}}}}}}}} };
\end{tikzpicture}
\ecc
\caption{Hierarchy~---~$\xb_{\sss 0}$ is the image of interest, $\xb_{\sss 1:T}$ are the latent images and $\yb$ is the measured image (blurred and noisy version of the true $\xb_{\sss 0}$). $\thetab$ contains the parameters of the observation (response and error), and its estimation is the core of the article. This graph already shows that if we know how to sample the $\xb_{\sss t}$ properly including the conditional independences encoded by this hierarchy, the difficulty of sampling $\thetab$ is greatly alleviated. \label{Fig:Hierarchy}}
\end{figure*}

\subsection{Full posterior}

We can then construct the joint pdf and the posterior. The latter is based on the likelihood~(\ref{Eq:Likelihood}) and the priors for the parameters~(\ref{Eq:PriorInstru}), (\ref{Eq:PriorNoiseOffset}), (\ref{Eq:PriorNoiseScale}), and the joint prior for the images (\ref{Eq:Forward})-(\ref{Eq:Backward}). Its construction relies on conditional independences encoded in the hierarchical model given in Fig.\,\ref{Fig:Hierarchy}. 
\beqn
\pi_{\sss 0:T}(\xb_{\sss 0:T},\thetab|\yb) 
	&\propto& \gamma_e^{P/2} \, \exp\,  -\gamma_e \norm{(\yb-m_e) - \Hb_\iotab \,\xb_{\sss 0}}^2 /2 \nonumber\\[0.1cm]
	& &  \gamma_e^{a_0-1} ~ \Exp{\, -b_0\,\gamma_e \,} \, \unbb_+(\gamma_e) \nonumber\\[0.1cm]
	& &  \Exp{\, -p_0 \, (m_e- m_0)^2 /2\, }  \nonumber\\[0.1cm]
	& &  \Uc(\iotab) \nonumber\\[0.0cm]
	& &  \pi_{\sss 0:T}(\xb_{\sss 0:T}) \label{Eq:FullPosterior}
\eeqn
Due to the intricate nature of this pdf, it is not possible to compute the estimations and uncertainties directly. To this end, an MCMC sampler is used, as shown below.

\section{Proposed sampler: a Gibbs scheme} \label{Sec:PosteriorSampler}

To explore the posterior, we resort to a Gibbs loop that splits the global sampling problem in easier sub-problems. More precisely, the conditional posterior of each unknown is sequentially sampled under its conditional density, in an iterative way. The samples form a Markov chain whose distribution converges to the posterior~\cite{Robert07,Brooks11}.

\brk The paper \cite{Murata23} is related to the work proposed here but there are two key differences. First, the estimation of an instrument parameter is considered in \cite{Murata23} but not the estimation of noise parameters (neither the offset nor the power), and second the Gibbs algorithm in \cite{Murata23} structures the alternation between the images and the instrument parameters but not between the latent variables themselves. \label{RQ:GibbsDDRM}
\erk

For each unknown, the conditional pdf given the other un\-knowns is needed. Each one is proportional to the pos\-te\-rior~(\ref{Eq:FullPosterior}) and hence only involves the factors including the considered unknown. 
Given the hierarchy in Fig.\,\ref{Fig:Hierarchy}, several simplifications arise, which both facilitate the theoretical calculations and reduces the computational load.
The conditional posteriors are now given. 
For notational simplicity $\bar\yb=\yb-m_e$.

\subsection{Image}

This section describes the sampling of the extended image $\xb_{\sss 0:T}$. Up to a factor, the pdf writes
\beqx 
	\Exp{ -  \frac{1}{2} \gamma_e \norm{ \bar\yb - \Hb_\iotab {\xb_{\sss 0}} }^2} ~\pi({\xb_{\sss 0:T}}) \,.
\eeqx
%
%
We resort to  G-DPS presented in~\cite{Giovannelli26}. It is itself a block-Gibbs sampler: it samples each $\xb_{\sss t}$ in turn under its conditional pdf $\pi_{\sss t|\star}(\xb_{\sss t}|\yb,\thetab,\xb_{\sss \star\backslash t})$ where $(t\I\star)$ is the time $t$ given all the other times (from $0$ to $T$) except $t$ and $(\star\backslash t)$ denotes the set of all times (from $0$ to $T$) except $t$.
The original idea of~\cite{Giovannelli26} is to play with both forward and backward pdfs. More specifically, the sampling is based on the posterior attached to the 
\bit
\item forward $\pi^{\sss +}_{\sss 0:T}(\xb_{\sss 0:T}|\yb,\thetab)$ for the latent variables $\xb_{\sss 1:T}$, and 
\item backward $\pi^{\sss -}_{\sss 0:T}(\xb_{\sss 0:T}|\yb,\thetab)$ for the image of interest $\xb_{\sss 0}$.
\eit
This idea is justified by the fact that the two joint priors $\pi^{\sss +}_{\sss 0:T}(\xb_{\sss 0:T})$ and $\pi^{\sss -}_{\sss 0:T}(\xb_{\sss 0:T})$ are similar thanks to the learning stage. So, we consider here that they are identical, then the convergence is considered as guaranteed.
Overall, the entire algorithm is both simple and efficient for three reasons.
\ben
\item It requires the sampling of Gaussians only (see also~\cite{Kamilov25}) 
\item All the covariances are diagonal be it in the Fourier domain ($t=0$) or in the spatial one ($t\neq0$).
\item In addition, means and variances are easy to compute, by FFT ($t=0$) or linear combination ($t\neq0$). 
\een
The main technical details are reported in Appendix and the full details are~\cite{Giovannelli26}.


\subsection{Noise parameter scale $\gamma_e$}

Up to a factor, the conditional posterior for $\gamma_e$ reads
\smallskip
\beqnx 
~~\gamma_e^{P/2} \Exp{-\frac{\gamma_e}{2} \norm{ \bar\yb - \Hb_\iotab \xb }^2 } \, \gamma_e^{a_0-1} \Exp{\, -b_0 \gamma_e \,} \, \unbb_+(\gamma_e) \\[0.1cm]
=~\gamma_e^{a_0+P/2-1} \, \Exp{-\gamma_e \pth{ b_0 + \frac{1}{2} \norm{ \bar\yb - \Hb_\iotab \xb }^2 }} \, \unbb_+(\gamma_e)
\eeqnx
and the advantage of a conjugacy becomes apparent at this point: the conditional posterior for $\gamma_e$ is in the same family as the prior, namely a Gamma pdf. The parameters are:
\beqx
\left\{
	\begin{array}{rcl}
	a &=& a_0 + P/2 \\[0.1cm]
	b &=& b_0 + \norm{ (\yb-m_e) - \Hb_\iotab \xb }^2 /\, 2
	\end{array}	
\right.
\eeqx
the sampling is then direct and efficient. 

\subsection{Noise parameter offset $m_e$}

The conditional posterior for $m_e$ clearly appears as:
\beqnx 
	&& \Exp{-\frac{ \gamma_e}{2}  \norm{ (\yb-m_e) - \Hb_\iotab \xb }^2 } \, \Exp{-p_0 \, (m_e- m_0)^2 /2} \\[0.1cm]
	& = & \Exp{- \frac{p}{2} (m_e-m)^2 }
\eeqnx
up to a factor, that is a Gauss pdf with precision and mean
\beqx
\left\{
	\begin{array}{rcl}
	p &=& p_0 + P \gamma_e \\[0.1cm]
	m &=& p\pmu \, ( p_0 m_0 + \gamma_e \, \unb\T(\yb-\Hb_\iotab\xb) )
	\end{array}	
\right.
\eeqx
and the sampling is also direct and efficient. At this point also, the advantage of a conjugacy is apparent (the prior and the conditional posterior are in the same family).

\subsection{Instrument parameter}

The conditional posterior for the instrument parameter $\iotab$ is also proportional to the joint posterior~(\ref{Eq:FullPosterior}):
\beqnx 
\Exp{-\frac{\gamma_e}{2}  \norm{ \bar\yb - \Hb_\iotab \, \xb }^2 } ~ \Uc(\iotab)
\eeqnx
This pdf is not an usual one and cannot be directly sampled. Among existing sampling algorithms~\cite{Robert07,Brooks11,Girolami11}, we resort to a Metropolis-Hasting step. Within this family of algorithms, several options are available (independent, random-walk,\dots). Here it is efficient to make use of random-walk Metropolis-Hasting with a Gauss excursion.

\section{Numerical assessment}\label{Sec:Simulations}\label{Sec:Results}


In order to demonstrate the feasibility and interest of the proposed Hyper-G-DPS, this Section proposes an experimental study. It relies on a toy problem based on the MNIST example set. The method has been implemented\footnote{within the Matlab environment on a standard PC with a 3.8~GHz CPU and 32~GB of RAM, with a GPU "NVIDIA GeForce RTX 3090".} and the information regarding the architecture and learning stage are given in \cite{Matlab23}. 
The ground-truth $\xb^\star$ is a sample of the learned prior (size $32\times 32$, gray level roughly in~$[0,1]$). The PSF is a Lorentz shape with width parameter $\iota^\star=0.9$ and regarding the noise $\sigma_e^\star=0.05$ and $m_e=0.1$. 
The ground-truth $\xb^\star$ and the measurement (blurred and noisy image) $\yb$, are shown in Fig.\,\ref{Fig:ResultsImages} (left and middle).
Here are some implementation details.
\bit
\item Regarding the scan order, the algorithm repeats this pattern: update observation parameters $v_e$ and $m_e$, then~$\iota$, followed by the images for $t=0$ up to $t=T$. 
\item The image $\xb_{0}$ is initialized to $\yb$. The $\xb_{\sss 1:T}$ are set to successive noisy versions through the forward model. 
%
%
The width $\iota$ and the error mean $m_e$ are initialized at random under their prior. Given the scan order and the Gibbs structure, there is no need to initialize~$v_e$. 
%
\item As the iterations proceed, the empirical average of the images is updated. The algorithm stops when the difference between successive updates is smaller than a threshold. 
\eit 
\textsl{Remark}~---~The algorithm has been run numerous times under identical and different scenarios, including variations in ground truth, noise level, PSF and initialisations. It has consistently exhibited both qualitative and quantitative behaviour.

\subsection{Results}

Fig.\,\ref{Fig:ParamChains} shows a typical result regarding the three unknown parameters. The chains exhibit standard behaviour: the distributions quickly stabilise and appear stationary after about only $300$ iterations (burn-in period).
From a qualitative view, Fig.\,\ref{Fig:ParamChains} shows that the estimated values are nearby the true values. The quantitative results are reported in Tab.\,\ref{Tab:ParamEstimate}. 

\bfig[htb]
\bcc
\btabu{c}
\includegraphics[width=3.7cm]{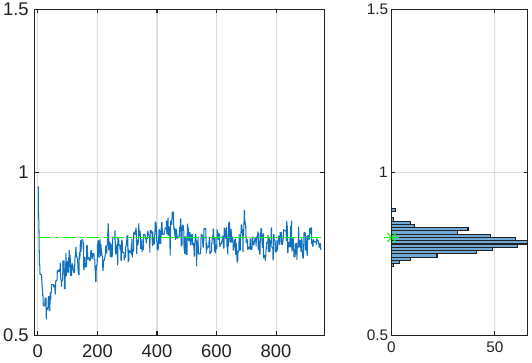}\\
\includegraphics[width=3.7cm]{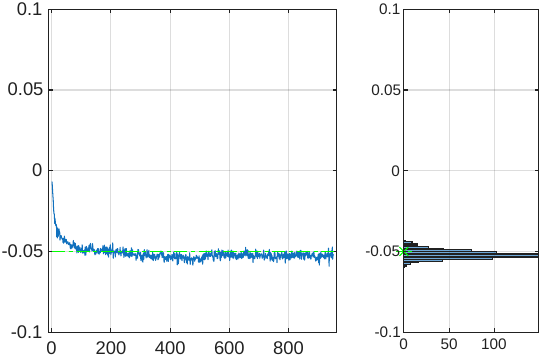}\\
\includegraphics[width=3.8cm]{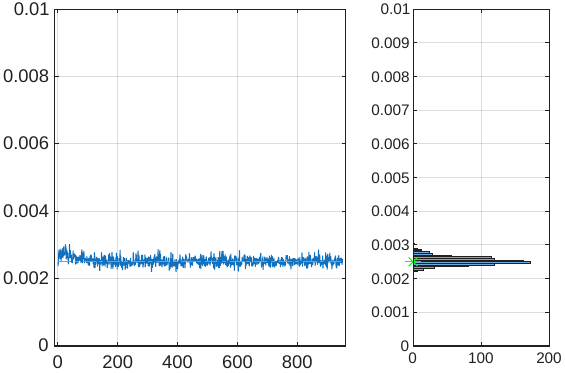}\\
\etabu
\ecc
\caption{Samples provided by the Gibbs algorithm as a function of iteration index (left) and as histograms (right) for the three unknown parameters from top to bottom: $\iota$, $m_e$ and $v_e$. They are samples of one dimensional marginal pdfs. The green lines\,/\,dots give the true value. See Fig.\,\ref{Fig:ParamClouds} for two dimensional joint-marginal posterior and Tab.\,\ref{Tab:ParamEstimate} for quantitative results. \label{Fig:ParamChains}} 
\efig


The proposed strategy provides optimal estimations (\eg Posterior Mean as the MMSE) and additionally coherent tools for uncertainty quantification based on posterior standard deviations. 
For each parameter, it is clear from Tab.\,\ref{Tab:ParamEstimate} that the true value lies within the interval centered on the estimate and of width two standard deviations.
%

\bfig[h]
\centering
\includegraphics[width=0.4\textwidth]{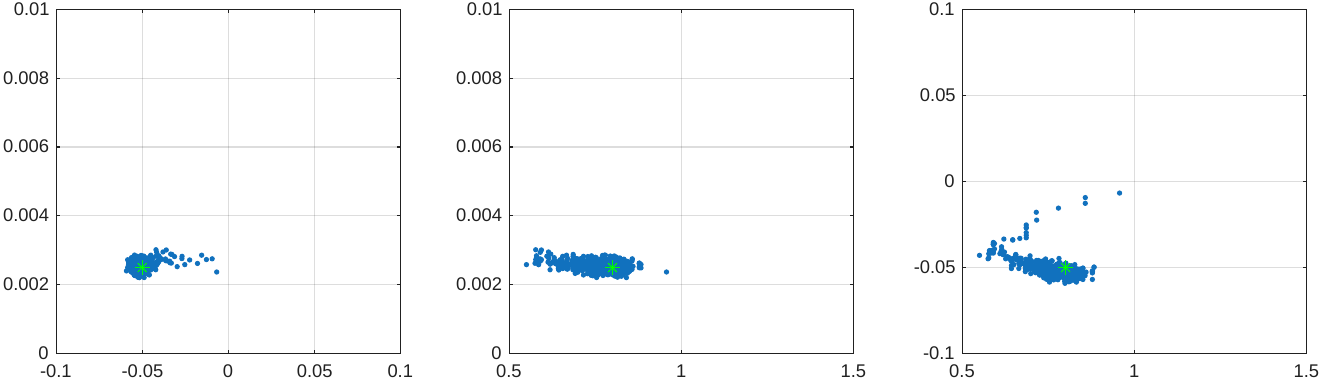} 
\caption{Point clouds for two dimensional marginals pdfs for the three unknown parameters: $\iota$, $m_e$ and $v_e$. From left to right: $(m_e,v_e)$,  $(\iota,v_e)$, and $(\iota,m_e)$. The samples are given in blue and the true values is given in green. See also Fig.\,\ref{Fig:ParamChains} for one dimensional plots and Tab.\,\ref{Tab:ParamEstimate} for quantitative assessment. \label{Fig:ParamClouds}}
\efig
%
%
\btabl[htb]
\bcc
\btabu{lllll}
            & $\iota$	& $m_e$		&  $v_e$ ($\times 10^3$)	\\[0.0cm] \hline
True        & 0.80  	& -0.050   	& 2.50		\\
Estimate    & 0.77  	& -0.051  	& 2.53		\\
Error       & 0.030		& ~0.0010 	& 0.026	\\
Error       & 3.8\% 	& ~2.1\% 	& 1.1\%		\\
PSD         & 0.053 	& ~0.0049 	& 0.122		\\
$\pm$ 2PSD  & ~~\MyCheck & ~~\MyCheck     & ~~\MyCheck  \\
\etabu
\ecc
\caption{Results for the three unknown parameters $\iota$, $m_e$ and $v_e$: true and estimated values (first and second row) then the error (third row). The Posterior Standard Deviation (PSD) is then given and the \MyCheck\, indicates that the true value does lie within the interval centered on the estimate and of width two~PSD. See also Figs.\,\ref{Fig:ParamChains} and~\ref{Fig:ParamClouds}. \label{Tab:ParamEstimate}}
\etabl

Finally, Fig.\,\ref{Fig:ResultsImages}-right yield the estimated image. The blur and the noise are significantly reduced in the resulting image (Fig.\,\ref{Fig:ResultsImages}-right) with respect to the measurement (Fig.\,\ref{Fig:ResultsImages}-middle) and it closely matches the original image (Fig.\,\ref{Fig:ResultsImages}-left). This result is confirmed by the cross-sections also shown in Fig.\,\ref{Fig:ResultsImages}. From Fig.\,\ref{Fig:ResultsUncertainty} it is clear that, for each pixel, the true value also lies within the interval centered on the estimate and of width two standard deviations.

\bfig[htb]
\btabu{rrr}
\includegraphics[width=2.3cm]{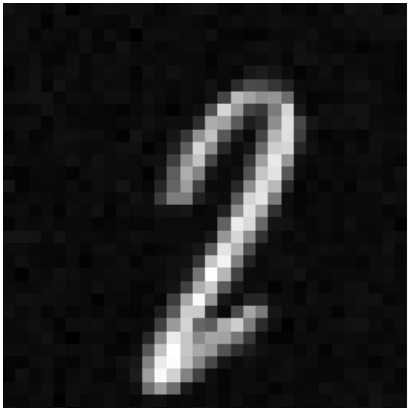} &
\includegraphics[width=2.3cm]{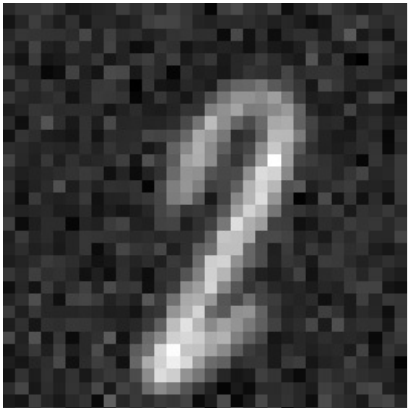}  &
\includegraphics[width=2.3cm]{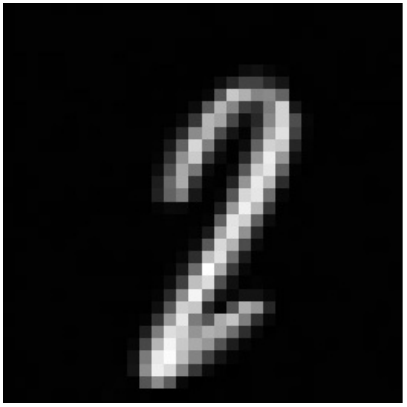}   \\
\includegraphics[width=2.5cm,height=1cm]{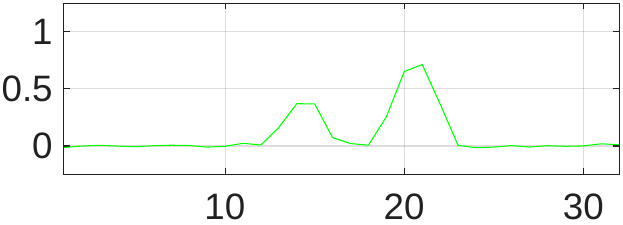} &
\includegraphics[width=2.5cm,height=1cm]{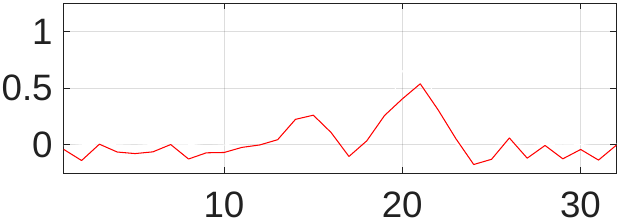}  &
\includegraphics[width=2.5cm,height=1cm]{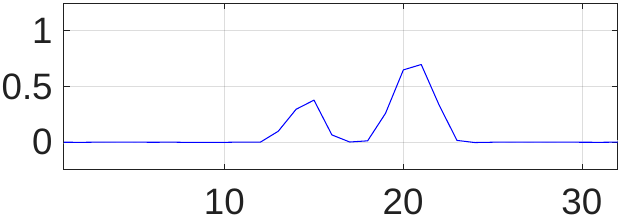}   \\
\etabu
\caption{Left to right: true image $\xb^\star$, measurements $\yb$ and estimated image~$\wh\xb$. The figure shows the images themselves (top) and cross-sections (bottom). \label{Fig:ResultsImages}}
\efig
\bfig[htb]
\bcc
\includegraphics[width=5cm]{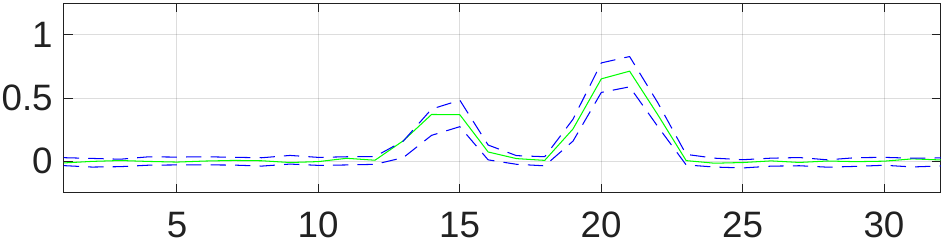}
\ecc
\caption{Cross-sections of $\xb^\star$ (plain green) and the \Quote{uncertainty} intervals (dashed blue) centered on the estimate and of width three standard deviations.} \label{Fig:ResultsUncertainty}
\efig

\pagebreak
\subsection{Efficiency, computation time and some comments} \label{Sec:SimulationsTime}

The algorithm produces $N$ samples of the images and the parameters, $\xb_{\sss 0:T}^{(n)}$ and $\thetab^{(n)}$ for $n=1,\dots N$ under the joint posterior pdf for $\xb_{\sss 0:T}$ and $\thetab$. Note that each iteration $n$ involves updating the three parameters $\thetab=\cro{\iota, m_e, v_e}$ and all the images $\xb_{\sss t}$ (for $t=0,\dots,T$).

As mentioned earlier, as the iterations progress, the empirical average of the images $\xb_{\sss 0}^{(n)}$ (which approximates the posterior mean) is updated. The iterations stop when the difference between successive updates becomes smaller than a threshold, here set to $10^{-2}$. The algorithm thus iterated $N=952$ times taking $62$~seconds. Most of the computations time (about 80\%) is due to the passage through the network.

A particular feature of the Hyper-G-DPS sampling scheme, inherited from G-DPS, is that, ultimately, each iteration $n$ (updating all the $\xb_{\sss t}$, for $t=0,1,\dots T$) requires only a single pass through the neural network (to update $\xb_{\sss 0}^{(n)}$). Therefore, scaling up to larger images does not appear to be an obstacle.

Another major practical advantage of Hyper-G-DSP, also inherited from G-DPS, is that it does not require the adjustment or tuning of any algorithm parameters (apart from the threshold that puts an end to the iterations), unlike many other algorithms, \eg~\cite{Chung24,Song23}.



\section{Conclusion}\label{Sec:Conclusion}

This paper deals with numerical methods for solving inverse problems when the observation model is linear with additive Gauss noise. The focus is on the delicate issue of estimating multiple parameters of the observation system: width of the point spread function and also mean and variance of noise\,/\,error. This issue has already been addressed in the literature and several solutions have been proposed, but not really (see Remark~\ref{RQ:GibbsDDRM}) in cases where the prior for the image is defined on the basis of a diffusion model. It is the specificity of the proposed method to estimate them in addition to the image of interest in that case. To this end,  our recent contribution~\cite{Giovannelli26} allows for proper handling of conditional distributions for images, thereby enabling the inclusion of the conditional posterior for parameters and posterior given these parameters. More precisely, a Gibbs loop splits the overall problem in far simpler sub-problems: iteratively sample each parameter and each image under its conditional posterior.
%


%

The simulation study focuses on parameter estimation issue and it is based on the MNIST example set. The proposed method provides accurate and coherent elements for uncertainty quantification, as well as accurate parameters estimation and image restoration. The numerical study also confirms the remarkable computational efficiency.

Conclusively, the paper addresses the crucial question of estimating instrument and noise parameters, in addition to the unknown image, in inverse problem based on a diffusion prior. It provides a novel solution referred to as Hyper-G-DPS, that is shown to be accurate and efficient.

%
%
To go further, it would be interesting to address model selection~\cite{Ando10,Ding18}, especially selection of a model for instrument an\,/\,or noise from a given list~\cite{Harroue24}.

\appendix

This appendix provides computational details regarding the Gibbs algorithm used to sample the extended image $\xb_{\sss 0:T}$. Our previous paper~\cite{Giovannelli26} introduced this algorithm and gives more details. It samples each $\xb_{\sss t}$ in turn under its conditional pdf $\pi_{\sss t|\star}(\xb_{\sss t}|\yb,\thetab,\xb_{\sss \star\backslash t})$ where $(t\I\star)$ is the time $t$ given all the other times (from $0$ to $T$) except $t$ and $(\star\backslash t)$ denotes the set of all times (from $0$ to $T$) except $t$.
The structure of these conditional pdfs relies on the hierarchy shown in Fig.\,\ref{Fig:Hierarchy}.

\medskip\indent\textit{1~-~Image of interest}~---~
Regarding $\xb_{\sss 0}$, it is sampled under 
\begin{align*}
\pi_{\sss 0|\star}&(\xb_{\sss 0}|\yb,\thetab,\xb_{\sss \star\backslash 0}) \\
	& \propto~  \pi^{\sss -}_{\sss 0|1} (\xb_{\sss 0} \I \xb_{\sss 1}) ~ f(\yb \I \xb_{0},\thetab)  \\
	& =~ \Nc( \xb_{\sss 0} \,; \mub_{\sss 1}(\xb_{\sss 1}) , \VarBack_{\sss 1} \Ib ) ~ \Nc(\yb \,;\, m_e+\Hb\xb_{\sss 0},v_e \Ib)
\end{align*}
which reveals a linear-Gauss problem and the Wie\-ner\,/\,Ti\-kho\-nov solution. So, the conditional posterior is Gauss with precision and expectation.
\beqx
\bca
    \Gammab_{\sss 0} &=~~ \Hb_\iotab\T \Hb_\iotab \,/\, v_e  +  \Ib \,/\,\VarBack_{\sss 1}  \\
    \varepsilonb_{\sss 0} &=~~ \Gammab_{\sss 0}\pmu\cro{ \Hb_\iotab\T (\yb-m_e) \,/\, v_e \,+\,  \mub_{\sss 1}(\xb_{1})\,/\,\VarBack_{\sss 1}}
\eca
\eeqx
Sampling is particularly effective in the Fourier plane: the components are independent and Gaussian, and their mean and variance are easily obtained by simple FFT~\cite{Orieux10}. 

\medskip\indent\textit{2.1~-~Latent images ($t\neq T$)}~---~The $\xb_{\sss t}$ are sampled under 
\begin{align*}
\pi_{\sss t|\star}&(\xb_{\sss t}|\yb,\xb_{\sss \star\backslash t}) \\
	& \propto~ \pi^{\sss +}_{\sss t|t-1} (\xb_{\sss t} \I \For_{\sss t-1}) ~ \pi^{\sss +}_{\sss t+1|t} (\xb_{\sss t+1} \I \xb_{\sss t})	\\
	& =~ \Nc (\xb_{\sss t}   \,;\, k_{\sss t} \, \xb_{\sss t-1} , \VarFor_{\sss t} \Ib  ) ~ \Nc (\xb_{\sss t+1}   \,;\, k_{\sss t+1} \, \xb_{\sss t} , \VarFor_{\sss t+1} \Ib  )
\end{align*}
also yields a Gauss pdf with precision $\gamma_{\sss t} \Ib$ and expectation~$\varepsilonb_{\sss t}$
\beqx
\bca
    \gamma_{\sss t} &=~~ 1 / \VarFor_{\sss t} \,+\,  k_{\sss t+1}^2 / \VarFor_{\sss t+1}  \\
    \varepsilonb_{\sss t} &=~~  \gamma_{\sss t}\pmu \pth{ k_{\sss t} \xb_{\sss t-1} / \VarFor_{\sss t} \,+\, k_{\sss t+1} \xb_{\sss t+1} / \VarFor_{\sss t+1} }
\eca
\eeqx

\medskip\indent\textit{2.2~-~Latent image ($t=T$)}~---~For the case of $\xb_{\sss T}$
\begin{align*}
\pi_{\sss T|\star}(\xb_{\sss T}|\yb,\xb_{\sss \star\backslash T}) 
	& =~ \pi^{\sss +}_{\sss T|T-1} (\For_{\sss T} \I \For_{\sss T-1}) \\
	& =~ \Nc (\For_{\sss T}   \,;\, k_{\sss T} \, \For_{\sss T-1} , \VarFor_{\sss T} \Ib  )
\end{align*}
\ie simply the last step in the forward process: a Gaussian with precision $\gamma_{\sss T} \Ib$ and expectation~$\varepsilonb_{\sss T}$%
\beqx
\bca
    \gamma_{\sss T} &=~~ 1/ \VarFor_{\sss T}   \\
    \varepsilonb_{\sss T} &=~~  k_{\sss T} \, \For_{\sss T-1}
\eca
\eeqx
\medskip

\section*{Acknowledgment}
The author warmly thanks Liam Moroy, Guillaume Bourmaud, Fr\'ed\'eric Champagnat, Marcelo Pereyra and Charlesquin Kemajou for their help.

This work is conducted within project \emph{PEPR Origins}, reference ANR-22-EXOR-0016, supported by the France~2030 plan managed by Agence Nationale de la Recherche. It also received financial support from the French government in the framework of the University of Bordeaux's France 2030 program \emph{RRI Origins}.

	\bibliographystyle{IEEEtran}
	\bibliography{biben,bibenabr,revuedef,revueabr,name,BaseBiblioJFG,BaseBiblioGPI,BaseBiblioAutre}

\edoc